\documentclass[10pt, a4paper]{article}
\usepackage{lrec}
\usepackage{multibib}
\newcites{languageresource}{Language Resources}
\usepackage{graphicx}
\usepackage{tabularx}
\usepackage{soul}

\usepackage{multicol}
\usepackage{multirow}
\usepackage{amsmath}
\usepackage{epstopdf}
\usepackage[latin1]{inputenc}
\usepackage{color}

\usepackage{hyperref}
\usepackage{xstring}

\DeclareSymbolFont{extraup}{U}{zavm}{m}{n}
\DeclareMathSymbol{\varheart}{\mathalpha}{extraup}{86}
\DeclareMathSymbol{\vardiamond}{\mathalpha}{extraup}{87}

\def\fndaff{$^\varheart$}
\def\sstaff{$^\spadesuit$}

\title{Chinese--Portuguese Machine Translation: \\A Study on Building Parallel Corpora from Comparable Texts}

\name{Siyou Liu\fndaff ~~~~ Longyue Wang\sstaff ~~~~ Chao-Hong Liu\sstaff}

\address{\\
\fndaff~School of Languages and Translation, Macao Polytechnic Institute, Macao S.A.R., China \\
\sstaff~ADAPT Centre, School of Computing, Dublin City University, Ireland \\
 \normalsize \tt violetal@ipm.edu.mo\\
 \normalsize \tt \{longyue.wang, chaohong.liu\}@adaptcentre.ie\\}

\abstract{
Although there are increasing and significant ties between China and Portuguese-speaking countries, there is not much parallel corpora in the Chinese--Portuguese language pair. Both languages are very populous, with 1.2 billion native Chinese speakers and 279 million native Portuguese speakers, the language pair, however, could be considered as low-resource in terms of available parallel corpora.
In this paper, we describe our methods to curate Chinese--Portuguese parallel corpora and evaluate their quality.
We extracted bilingual data from Macao government websites and proposed a hierarchical strategy to build a large parallel corpus. Experiments are conducted on existing and our corpora using both Phrased-Based Machine Translation (PBMT) and the state-of-the-art Neural Machine Translation (NMT) models. The results of this work can be used as a benchmark for future Chinese--Portuguese MT systems. The approach we used in this paper also shows a good example on how to boost performance of MT systems for low-resource language pairs.
\\\newline  }

\begin{document}

\maketitleabstract

\section{Introduction}


Chinese and Portuguese are widely used by a large amount of people in the world. With the development of economic globalization, communications between Chinese and Portuguese-speaking countries are increasing in a fast path. Translation services between these two languages is becoming more and more demanding. However, Chinese and Portuguese belong to distinct language families (Sino-Tibetan and Romance, respectively) and only a relative much smaller proportion of people have bilingual proficiency of the language pair. Therefore, the use of Chinese--Portuguese MT systems to provide auxiliary translation services between the two sides is highly demanded.

Pivot-based machine translation is a commonly used method when large quantities of parallel data are not readily available for some language pairs. \newcite{N07-1061}, \newcite{wu2007pivot}, \newcite{bertoldi2008phrase} investigated phrase-level, sentence-level and system-level pivot strategies for low resource translation in SMT.
A pivot language, which is usually English, can bridge the source and target languages and make translation possible. However, the domains of these two are often different and thus results in low performance and even ambiguities.

A few researchers have investigated how to improve the Chinese--Portuguese MT by incorporating linguistic knowledge into the systems.
For instance, \newcite{wong2010pct} proposed a hybrid MT system combining rule-based and example-based components. \newcite{Oliveira2010} explored Constraint Synchronous Grammar parsing for SMT. \newcite{lu2014analysis} and \newcite{liu2016analysis} focused on specific linguistic phenomena (i.e. present articles and temporal adverbials) in translation.
Although NMT has been rapidly developed in recent years \cite{D13-1176,sutskever2014sequence,bahdanau2015neural,Tu:2016:ACL}, Chinese--Portuguese MT has not received much attention using NMT because training data are not readily enough. Therefore the performance is still low using these state-of-the-art approaches.

To date, there are only a few Chinese--Portuguese parallel corpora available\footnote{\url{http://opus.nlpl.eu}.} \cite{DBLP:conf/lrec/Tiedemann12}. OpenSubtitles2018\footnote{\url{http://opus.nlpl.eu/OpenSubtitles2018.php}.} \cite{DBLP:conf/lrec/LisonT16} has released 6.7 millions Chinese--Portuguese sentence pairs, which are extracted from movie subtitles. These sentences are usually short and simple as most of them are transcripts of conversations in movies, therefore they alone are not suitable to train general-domain MT systems. News-Commentary11\footnote{\url{http://www.casmacat.eu/corpus/news-commentary.html}.} contains data in newswire domain. However, there are only 21.8 thousands of sentence pairs and is thus not sufficient to train robust MT models.

\begin{figure*}[h]
\graphicspath{ {figures/} }
\begin{center}
\includegraphics[scale=0.42]{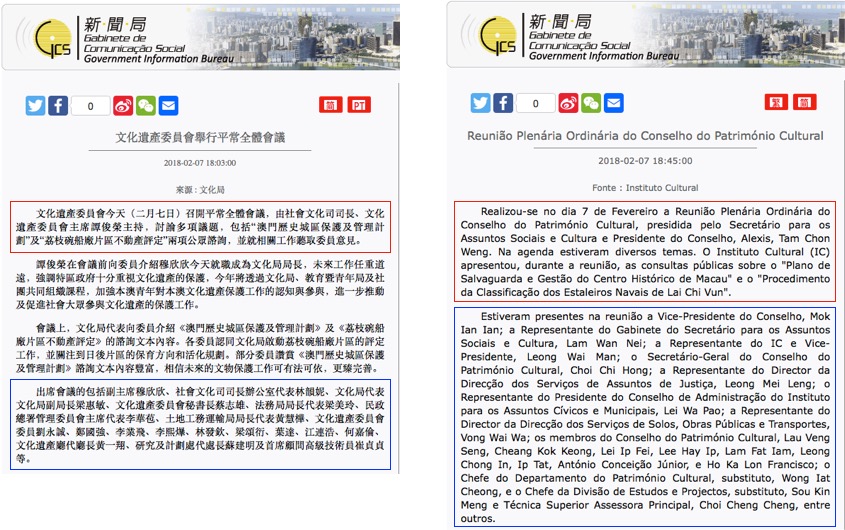} 
\caption{An example from \textit{The Government Information Bureau} website. The texts in the boxes with the same color are parallel data.}
\label{fig.1}
\end{center}
\end{figure*}

To alleviate data scarcity problem, we extracted bilingual data from Macao government websites.\footnote{Macao is a multi-cultural society in which both Mandarin Chinese and Portuguese are recognized as official languages.}  Macao government documents, as requested by law, are written and archived in both languages.  Domains contained in these documents include international communication, trade exchanges, technological cooperation, etc.  In order to build a high-quality parallel corpus, we propose a hierarchical strategy to deal with document-level, paragraph-level and sentence-level alignment. In total more than 800 thousands of Chinese--Portuguese sentence pairs in newswire, law and travelling domains, among others, are curated. Finally, we conducted experiments on Chinese--Portuguese machine translation tasks using both OpenSubtitles2018 and the curated corpus to evaluate the quality of the corpus.  The experimental results show that the performance of Chinese--Portuguese MT has significantly improved and outperforms the results using pivot-based method.
The contributions of this paper are listed as follows:
\begin{itemize}
\item We propose a hierarchical alignment approach to build a large and high-quality general-domain corpus, which in total contains more than 800 thousands of Chinese--Portuguese sentence pairs;
\item We evaluate the quality of the curated corpus with MT performance of Chinese--Portuguese MT systems trained on the two large corpora (i.e., OpenSubtitles2018 and ours); 
\item We investigate both SMT and NMT models and compare them using pivot-based method. The experimental results can be used as Chinese--Portuguese MT benchmark for future work.
\end{itemize}

The rest of the paper is organized as follows. In Section 2, we introduce our approach to build the Chinese--Portuguese corpus. The experimental results of MT tasks, which are used to evaluate the quality of corpus, are reported in Section 3. Analyses of out-of-vocabulary (OOV) and pivot-based MT are given in Section~4. Finally, Section 5 presents our conclusions and future work.




\section{Building a Chinese--Portuguese Corpus}

There are a number of Macao websites (e.g. \textit{The Government Information Bureau} website,\footnote{\url{http://www.gcs.gov.mo}.} \textit{The Macao Law},\footnote{\url{http://www.macaolaw.gov.mo}.} and \textit{The Government Printing Bureau}\footnote{\url{http://www.io.gov.mo}.}) containing bilingual resources. \textit{The Government Information Bureau} website, for example, contains more than 100 thousands of local news articles from the year 2000 to date and more than 80\% of the articles are written in both Portuguese and Chinese languages. As shown in Figure \ref{fig.1}, the same news is written in a ``comparable'' way.\footnote{A Comparable Corpus is a collection of ``similar'' texts in different languages or in different presentation forms of a language.} Although not completely aligned, there are still paragraphs can be aligned to each other. Therefore, these are still good resources to curate parallel corpora. We crawl all similar websites in Macao government website list\footnote{\url{ https://www.gov.mo/en/about-government/departments-and-agencies/}.} and we only use \textit{The Government Information Bureau} website for detailed discussions in the rest of the paper.

\begin{figure*}
\graphicspath{ {figures/} }
\begin{center}
\includegraphics[scale=0.56]{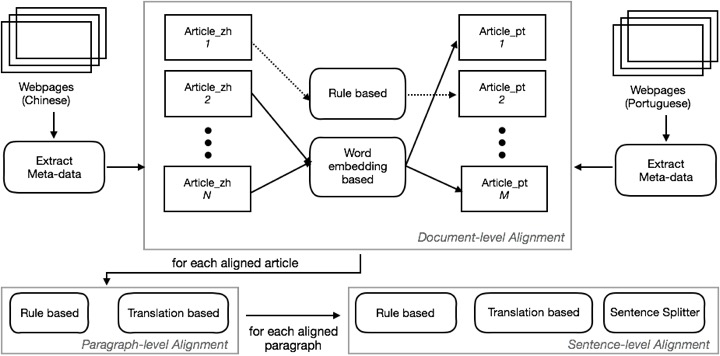} 
\caption{The workflow of building our Chinese--Portuguese corpus.}
\label{fig.2}
\end{center}
\end{figure*} 


As shown in Figure~\ref{fig.2}, we develop an end-to-end system to automatically build our parallel corpus from bilingual websites. However, there are still some challenges: 1) how to identify parallel/comparable news articles (bilingual document alignment tasks); 2) bilingual news articles are not direct translations to each other but written separately by Chinese authors and Portuguese authors of the same story.  Therefore these articles are mostly comparable rather than parallel texts (paragraph alignment tasks); 3) Chinese texts are usually written in chronicle style, while its corresponding Portuguese texts written in several sentences. Thus, these sentences are not one-to-one aligned (sentence alignment tasks). In order to obtain high-quality sentence pairs, we propose a hierarchical alignment strategy including document-level, paragraph-level and sentence-level alignment. At each level, we employ hybrid alignment approaches such as rule-based and translation-based. The architecture of our method can be described in a pipeline as follows: 
\begin{itemize}
\item [(1)] We crawl all accessible web pages from each website and then extract meta-data (e.g. title, author, date and content) using HTML tags;
\item [(2)] For document-level alignment, we first use URLs to align portions in articles. For other articles in which the heuristic rules do not apply, we employ a document alignment algorithm that calculates semantic similarity based on word embedding;
\item [(3)] For each aligned articles, we align the paragraphs with a simple but effective method: if the number of paragraphs in the two aligned articles is equal, we align all paragraphs one by one in the same order. Otherwise, we employ translation-based alignment algorithm to find parallel paragraphs;
\item [(4)] Within each aligned paragraph, we firstly use sentence boundary detection toolkit to split the paragraph into sentences and then do sentence-level alignment, which is similar to the process used in paragraph-level alignment. 
\end{itemize}



\subsection{Rule-based Alignment}

We found that around 10\% of web pages have already been aligned by URL links. As shown in Figure~\ref{fig.1}, there is usually a language switch button on the top of a web page which can be used to extract its corresponding page in the other language. Thus, we extract these useful links and align documents based on this heuristic rule. For other web pages the rule does not apply, we employ word embedding based alignment approach as described in Section~2.2.


At paragraph- and sentence-level alignment, we propose a simple but effective rules: 1) in each article/paragraph, we count the number of paragraphs/sentences on two sides. If the numbers are equal to each other, we align these paragraphs/sentences one by one in the same order. Otherwise, for example, the number of the source-side sentences is 10 while the number of the target-side is 12, we use translation-based alignment method (in Section~2.3) instead.


Using rule-based methods, we can easily obtain a reliable parallel sub-corpus. Although the corpus size is relatively small, these data are in the same domain. Thus the corpus can still be used to train a MT system for further steps. For instance, we could train a SMT system on the sub-corpus in newswire domain and use the system to translate sentences for translation based alignment method.

\begin{table*}[h]
\centering
\renewcommand\arraystretch{1.3}
\begin{tabular}{c | c|c cc c cc c cc}
\multirow{2}{*}{\bf Corpus} & \multirow{2}{*}{\bf Set}	&	\multirow{2}{*}{$|S|$}	&	\multicolumn{2}{c}{$|W|$} & &	\multicolumn{2}{c}{$|V|$} & & 	\multicolumn{2}{c}{$|L|$} \\
\cline{4-5}\cline{7-8}\cline{10-11}
			&	&	&	Zh	&	Pt	&	&	Zh	&	Pt	&	&	Zh	&	Pt\\
\hline

\multirow{3}{*}{\bf Opensub} & Train	&	6.51M & 48.14M & 53.45M & & 0.40M & 0.26M & & 7.39 & 8.21\\
 & Tune	&	2.07K & 15.43K & 18.22K & & 3.16K & 2.92K & & 7.46 & 8.81\\
 & Test	&	2.16K & 1243K & 17.49K & & 2.69K & 2.94K & & 5.74 & 8.08\\\hline

\multirow{3}{*}{\bf Our Corpus} & Train	&	0.84M & 17.02M & 22.49M & & 0.21M & 0.23M & & 20.36 & 26.90\\
 & Tune	&	1.00K & 19.90K & 30.83K & & 3.88K & 4.20K & & 19.90 & 30.83\\
 & Test	&	1.00K & 26.79K & 38.60K & & 4.68K & 4.99K & & 26.79 & 38.60\\
\end{tabular}
\caption{Number of sentences ($|S|$), words ($|W|$), vocabulary ($|V|$), and averaged sentence length ($|L|$) in the corpus. K stands for thousands and M for millions.}
\label{table.1}
\end{table*}

\subsection{Word Embedding based Alignment}

To align articles/documents, we consider the problem as cross-lingual document alignment task~\cite{wang2012improvement}. We employ a document alignment approach using word embedding~\cite{lohar2016fada}: 1) we initially construct a pseudo-query from a source-language document; 2) and then represent both the target-language documents and the pseudo-query as word vectors to find the average similarity measure between them; 3) finally the word vector based similarity is combined with the term-overlap-based similarity. 

The Vector Space Model (VSM) \cite{salton1975vector} is one of the overlap based methods. Each document is represented as a vector of terms. The $i$th document $D_i$ in target-side is represented as a vector $D_i = [w_{1,i}, w_{2,i}, ... w_{k,i}]$, in which $k$ is the size of the term vocabulary. Here we employ the cosine distance to calculate the similarity between two document vectors:
\begin{eqnarray}
\textit{sim}(d_i, d_j) = {\sum_{k=1}^{N} w_{i,k} \cdot w_{j,k} } { \sqrt {\sum_{k=1}^{N} w_{i,k} } \cdot {\sqrt {\sum_{k=1}^{N} w_{j,k} } } }
\end{eqnarray}
where $N$ is the number of terms in a vector, and $w_{i,k}$ and $w_{j,k}$ represent the weight of the $i\textit{th}$/$j\textit{th}$ term in $D_{i}$/$D_{j}$ respectively. Technically, the distance between documents in VSM is calculated by comparing the deviation of angles between vectors. A Boolean Retrieval Model sets a term weight to be either 0 or 1, while an alternative solution is to calculate the term weights according to the appearance of a term within the document collection. To calculate the term weights according to the appearance of a term within the document collection, we use term frequency-inverse document frequency (\textit{TF-IDF}) \cite{ramos2003using} as the term-weighting model. 


We also use the vector embedding of words and incorporate them with the VSM approach as mentioned above to estimate the semantic similarity between the source-language and the target-language documents. In practice, we indexed articles in both sides and then generate a query for each source-side article. Then we use a Chinese--Portuguese SMT system (training data are extracted using the method in Section~2.1) to obtain translated queries.


\subsection{Translation based Alignment}

Through exploring various sentence-alignment methods (e.g. length-based, dictionary-based), we found that translation based alignment is a robust approach especially for comparable data \cite{sennrich2010mt,DBLP:conf/nodalida/SennrichV11}. The idea is to use machine translated text and BLEU as a similarity score to find reliable alignments which are used as anchor points. The gaps between these anchor points are then filled using BLEU-based and length-based heuristics.

We use this method to align unaligned paragraphs and sentences. A Chinese--Portuguese SMT system (training data are extracted using the method in Section~2.1) is used to obtain translated paragraphs/sentences. 

\subsection{Machine Translation}
MT is a sequence-to-sequence prediction task, which aims to find for the source language sentence the most probable target language sentence that shares the same meaning. We can formulate SMT as: $\hat{\mathbf{y}} = {\arg\max}_{\mathbf{y}}\,p(\mathbf{y}|\mathbf{x})$ \cite{brown1993mathematics}, where $\mathbf{x}$ and $\mathbf{y}$ are sentences in source and target sides, respectively. $\hat{\mathbf{y}}$ denotes the translation output with the highest translation probability. $p(\mathbf{y}|\mathbf{x})$ is usually decomposed using the log-linear model:
\begin{equation}
\hat{\mathbf{y}} = \underset{\mathbf{y}}{\arg\max}\frac{\exp(\sum_{i=1}^{I}\lambda_i h_i(\mathbf{x},\mathbf{y}))}{\sum_{\mathbf{y}^\prime}\exp(\sum_{i=1}^{I}\lambda_i h_i(\mathbf{x},\mathbf{y}^\prime))}
\end{equation}
where $h_i(\cdot)$ indicates the translation feature and $\lambda_i$ is its corresponding optimal weight, which is learned by maximizing with a development set. $I$ indicates the total feature number. We employ phrase-based SMT in our experiments.

NMT is a new paradigm for MT in which a large neural network is trained to maximize the conditional likelihood on the bilingual training data. It directly models the probability of translation from the source sentence to the target sentence word by word \cite{D13-1176}:
\begin{equation}
P(\mathbf{y}|\mathbf{x})= \prod_{j=0}^{N} P (y_j | y_{<j}, \mathbf{x})
\end{equation}
in which given $\mathbf{x}$ and previous target translations $y_{<j}$ ($y_1, ..., y_{j-1}$), we need to compute the probability of the next word $y_j$ ($j \in \{1, ..., N\}$). We employ both RNNsearch \cite{sutskever2014sequence} and Transformer \cite{vaswani2017attention} architectures in our experiments.



\section{Experiments}

\subsection{Data}

The general domain parallel corpus is built using the approaches introduced in Section~2. We randomly sampled 1000 sentences and found that the alignment accuracy is over 94\%, indicates the corpus could be used for MT training. Regarding the document-level alignment, we use FaDA toolkit\footnote{\url{https://github.com/gdebasis/cldocalign}.} and for the translation based alignment, we employ Bleualign\footnote{\url{https://github.com/rsennrich/Bleualign}.}. To pre-process the raw data, we apply a series of procedures~\cite{wang2016lrec2} including: full/half-width conversion, Unicode conversation, simplified/traditional Chinese conversion, punctuation normalization, English/Chinese tokenization and sentence boundary detection, letter casing and word stemming, etc. For Portuguese tokenization and sentence splitting, we use Moses toolkit\footnote{\url{https://github.com/moses-smt/mosesdecoder}.}.
We randomly select one thousand (for the curated corpus) and approximately two thousands (for OpenSubtitles2018) of sentences as development and test sets; the numbers of sentences selected reflect the average sentence lengths of the two corpora. Table~\ref{table.1} lists the statistics of our corpus and OpenSubtitles2018 (\textit{Opensub}). 
Movie subtitle corpus is much larger than ours, however, its sentences are mostly simple and short.



\subsection{Setup}

We carry out our experiments on both Chinese-to-Portuguese and Portuguese-to-Chinese translation directions. We investigate various MT models: phrase-based SMT, RNNsearch NMT and Transformer NMT. We used case-insensitive 4-gram NIST BLEU metrics \cite{Papineni:2002} for evaluation, and {\em sign-test} \cite{Collins05} for statistical significance test.

\paragraph{SMT}
We employ Moses~\cite{Koehn:ACL:2007} to build phrase-based SMT model. The $5$-gram language model are trained using the SRI Language Toolkit~\cite{Stolcke:2002:CSLP}. To obtain word alignment, we run GIZA++~\cite{Och:2003} on the training data together with News-Commentary11 corpora. We use minimum error rate training~\cite{Och:2003b} to optimize the feature weights. The maximum length of sentences is set as 80.

\paragraph{RNNsearch}
We use our re-implemented attention-based NMT system, which incorporates dropout \cite{hinton2012improving} on the output layer and improves the attention model by feeding the most recently generated word. We limited the source and target vocabularies to the most frequent 50K and 30K words in Chinese and Portuguese, covering more than 97\% of the words in both languages. Each model was trained on sentences of lengths up to 80 words with early stopping. Mini-bataches were shuffled during processing with a mini-batch size of 80. The word-embedding dimension was 620 and the hidden layer size was 1,000. We trained for 20 epochs using Adadelta~\cite{zeiler2012adadelta}, and selected the model that yields best performance on the validation set.

\paragraph{Transformer}
We use our re-implemented Transformer NMT system. Most parameters are same as RNNsearch model except that 1) the encoder and decoder are both composed of a stack of 6 identical layers; 2) the hidden layer size is 512; 3) the batch size is 4096 tokens and 4) we use two GPUs for training.

\begin{table}[t]
\renewcommand\arraystretch{1.3}
\begin{center}
\begin{tabular}{c|c|c c}
 \bf Corpus & \bf System & \bf Dev & \bf Test\\
 \hline
 \multirow{3}{*}{Opensub} & SMT & 15.05 & 6.73 \\
 & RNNsearch & 13.37 & 13.34 \\
 & Transformer & 17.00 & 17.43 \\
 \hline
 \multirow{3}{*}{Ours} & SMT & 33.78 & 27.42\\
 & RNNsearch & 31.36 & 24.74\\
 & Transformer & 32.55 & 25.11\\
\end{tabular}
\caption{Results of Chinese-to-Portuguese translation.}
 \end{center}
 \label{table.2}
\end{table}

\subsection{Results}

Table~2 and Table~3 show the performances of different MT systems on Chinese-to-Portuguese and Portuguese-to-Chinese, respectively. 

\paragraph{Chinese-to-Portuguese Translation}
On \textit{Opensub}, SMT system only obtain 6.73 in BLEU score. NMT systems outperform it by 6--10 BLEU scores. We think SMT model is weak in translating informal domain (e.g. spoken domain) data, while distributed word representations can facilitate the computation of semantic distance. With our corpus, the SMT system can achieve 27.42 in BLEU while the best NMT model (Transformer) is around 2 points lower than SMT model. It is not surprising that the performance of NMT models have not surpassed that of traditional SMT (25.11 vs. 27.42). There are three main reasons: 1) the vocabulary size of \textit{Opensub} is very large, which results in a lot of out-of-vocabulary words (OOVs) in NMT training; 2) \textit{Opensub} is of small scale and the corpus is not big enough for NMT models to learn some general translation knowledge; 3) the sentences in our corpus is much longer than that in \textit{Opensub}. We will discuss these in Section~4. Furthermore, Transformer is the best one among NMT models in both corpora. The Transformer model is usually better than RNNsearch even for low-resource MT.

Generally, the BLEU scores using \textit{Opensub} (i.e. OpenSubtitle2018 corpus) are much lower than the scores with our corpus. For example, the performance on \textit{Opensub} is 6--17 in BLEU while it is 24--27 points on our data. Because sentences in movie subtitles are usually compact (i.e. short sentences with rich information) and contain multiple expressions, it results in a number of problems such as ambiguities for MT. 


\begin{table}[t]
\renewcommand\arraystretch{1.3}
\begin{center}
\begin{tabular}{c|c|c c}
 \bf Corpus & \bf System & \bf Dev & \bf Test\\
 \hline
 \multirow{3}{*}{Opensub} & SMT & 11.62 & 5.11\\
 & RNNsearch & 10.71 & 11.76\\
 & Transformer & 12.78 & 12.43 \\
 \hline
 \multirow{3}{*}{Ours} & SMT & 26.14 & 19.29\\
 & RNNsearch & 25.13 & 17.82\\
 & Transformer & 26.23 & 18.68\\
\end{tabular}
\caption{Results of Portuguese-to-Chinese translation.}
 \end{center}
 \label{table.3}
\end{table}

\paragraph{Portuguese-to-Chinese Translation}
As shown in Table~3, the translation performances of different systems on inverse direction are similar to those in Table~2. For example, NMT models still perform better than SMT model on \textit{Opensub} while worse on our general domain corpus. However, with our corpus, the performance of Transformer is close to that of SMT (i.e. -0.61 in BLEU). 

The BLEU scores in the Chinese-to-Portuguese direction are much higher than those in the inverse direction. Taking the performance on our data for instance, Chinese-to-Portuguese MT systems can usually achieve around 25 in BLEU whereas the systems for the inverse direction can only obtain about 18. It indicates that generating fluent and adequate Chinese translations is a more difficult task to MT systems.


The performance of Chinese--Portuguese machine translation is relatively lower; Chinese--English systems can usually achieve 36--40 in BLEU on NIST test sets. One of the reasons is that the Chinese--Portuguese language pair is low-resource. Another reason is the great distance between Chinese and Portuguese in terms of language families; there are extensive differences in syntax, semantics and discourse structures.

\section{Analysis}

In this section, we first discuss the OOV problem observed in the experimental results on our corpus (as described in Section~3.3). We also compare training directly using the parallel corpus we curated with the pivot-based method, which is a common approach for low-resource MT (as described in Section 1).

\subsection{Out-of-Vocabulary}

As shown in Table~1, vocabulary size is very big on both corpora. However, NMT models typically operate with a fixed vocabulary, which results in the OOV problem. This might contribute to the under-performance of NMT models compared SMT as observed in our experimental results.

Joint byte-pair encoding (BPE) \cite{P16-1162} is a simpler and more effective method to handle the OOV problem. It encodes rare and unknown words as sequences of subword units.  We use the BPE toolkit\footnote{\url{https://github.com/rsennrich/subword-nmt}.} to process our corpus and train an NMT system on this processed data. The procedures are as follows.  The Portuguese and Chinese data are first pre-processed using the same method introduced in Section~3.1. We then train single BPE models on tokenized/segmented both Portuguese and Chinese sides.  Finally, we use BPE models to segment the sentences into subword units.

After 59,500 joint BPE operations, the network vocabulary sizes are reduced to 67K and 59K for Chinese and Portuguese sides, respectively. Compared with the original vocabulary sizes (i.e. 210K and 230K), BPE method has significantly alleviate the problem of OOV.  We train a new Chinese-to-Portuguese NMT model with Transformer on BPE-based data. As shown in Table~4, the NMT model trained on BPE data increase 1.44 points in BLEU compared that without BPE.  The performance is getting closer to that of SMT, which shows using BPE with subword units does deal with OOV problem to some extent.

\begin{table}[!h]
\renewcommand\arraystretch{1.3}
\begin{center}
\begin{tabular}{c|c c}
 \bf System & \bf Dev & \bf Test\\
 \hline
 SMT & 33.78 & 27.42 \\
 Transformer & 32.55 & 25.11 \\
 Transformer + BPE & 33.96 & 26.55 \\
\end{tabular}
\caption{Comparisons of results using SMT and NMT trained with/without BPE (Chinese-to-Portuguese).}
 \end{center}
 \label{table.5}
\end{table}

\subsection{Pivot-based MT}

As discussed in Section~1, pivot method is commonly used for low-resource MT.  To show that increasing parallel data is still essential to improve low-resource MT, we also compare MT models trained with parallel corpus of direct translation pair, with the pivot-based models.

We build four Transformer NMT models on large Chinese--English\footnote{\url{http://www.statmt.org/wmt17.}} and English--Portuguese\footnote{\url{http://www.statmt.org/europarl.}} parallel corpora: Chinese-to-English, English-to-Portuguese, English-to-Chinese, Portuguese-to-English NMT models.  Taking English as the pivot language, we firstly use Chinese-to-English model to translate the Chinese input into English. Secondly, we use English-to-Portuguese model to translate the machine translated English sentences into Chinese output.  For the Portuguese-to-Chinese translation direction, the experiments are administered in the same manner.

As shown in Table~5, the pivot-based systems perform poor than those trained on direct parallel corpora. For instance, Portuguese-to-English-to-Chinese (``PT-EN-ZH'') system obtains only 11.29 in BLEU on our test set which is 7.39 points lower than our Transformer model. It indicates that increasing the amount of parallel data does help improve low-resource MT systems. 

\begin{table}[h]
\renewcommand\arraystretch{1.3}
\begin{center}
\begin{tabular}{c|c|c c}
 \bf Direction & \bf Corpus & \bf Dev & \bf Test\\
 \hline
 \multirow{2}{*}{ZH-EN-PT} & Opensub & 8.23 & 10.52\\
 & Ours & 15.38 & 14.60\\
 \hline
 \multirow{2}{*}{PT-EN-ZH} & Opensub & 8.88 & 10.33\\
 & Ours & 12.31 & 11.29\\
\end{tabular}
\caption{Results of pivot-based translation.}
 \end{center}
 \label{table.5}
\end{table}


\section{Conclusions and Future Work}

In this paper we described our methods to build a large Chinese--Portuguese corpus. Despite both Chinese and Portuguese are populous languages, the language pair itself could be considered as low-resource. Therefore the same technologies could be used to improve machine translation quality in other low-resource language pairs.
We conduct experiments on existing and the curated corpora, and compare the performance of different MT models using these corpora. This results of this work can be used by Chinese--Portuguese MT research for comparison purposes.

In the future, we will investigate other approaches such as universal low-resource NMT \cite{jiatao} and discourse-aware approaches \cite{wang2018aaai,D17-1300,wang2016naacl} for Chinese--Portuguese MT task. Furthermore, we will keep exploring simple yet effective methods to build larger and domain-specific Chinese--Portuguese parallel corpora to further improve MT performance in this language pair.


%

%
%
%
%

\section{Acknowledgements}

The ADAPT Centre for Digital Content Technology is funded under the SFI Research Centres Programme (Grant No. 13/RC/2106) and is co-funded under the European Regional Development Fund. This project has partially received funding from the European Union's Horizon 2020 Research and Innovation programme under the Marie Sk{\l}odowska-Curie Actions (Grant No. 734211).

\section{References}
\label{main:ref}

\bibliographystyle{lrec}
\bibliography{lrec}


\end{document}